# Cross-Architecture Knowledge Distillation (KD) for Retinal Fundus Image Anomaly Detection on NVIDIA Jetson Nano

E6692.2025.Spring.AABY.report.by2385.aa5479

Berk Yilmaz(by2385), Aniruddh Aiyengar (aa5479)
*Columbia University*

## Abstract

*Early and accurate identification of retinal ailments is crucial for averting ocular decline however access to dependable diagnostic devices is not often available in low-resourced settings. This project proposes to solve that by developing a lightweight, edge-device deployable disease classifier using cross-architecture knowledge distilling. We first train a high-capacity vision transformer (ViT) teacher model, pre-trained using I-JEPA self-supervised learning, to classify fundus images into four classes: Normal, Diabetic Retinopathy, Glaucoma, and Cataract. We kept an internet of things (IoT) focus when compressing to a cnn-based student model for deployment in resource limited conditions, such as the NVIDIA Jetson Nano. This was accomplished using a novel framework which included a Partitioned Cross-Attention (PCA) projector, a Group-Wise Linear (GL) projector, and a multi-view robust training method. The teacher model has 97.4 percent more parameters than the student model with it achieving 89 percent classification with a roughly 93 percent retention of the teacher model's diagnostic performance. The retention of clinical classification behavior supports our methods initial aim: compression of the ViT while retaining accuracy. Our work serves as an example of a scalable, AI-driven triage solution for retinal disorders in under-resourced areas.*

## 1. Introduction

Automated retinal fundus image analysis enables early detection and diagnosis of ocular diseases to help prevent vision loss and initiate timely interventions. Diabetic Retinopathy (DR), Glaucoma, and Cataracts can all be detected via retinal imaging with careful consideration of the fundus image. While this traditionally involves specialized equipment and trained ophthalmologists, we can leverage the recent attention to deep learning and specifically, the Vision Transformer (ViT) models to replicate this diagnostic procedure with tremendous accuracy. Our teacher ViT model achieved an accuracy of 92.87% in four classifications, Normal, Glaucoma, DR, and Cataract. Unfortunately, these models have deployment challenges in resource-limited clinical settings due to their computationally large sizes.

However, deploying complex transformer-based models on limited hardware creates significant technical challenges: the memory size of ViTs exceeds the capacity of the Nano, inference latency is too long to be clinically useful, and power efficiency for continuous use in limited-resources locations is divorced from applicable processing efficiency metrics.

To address these challenges, we propose a cross-architecture knowledge distillation framework that efficiently transfers diagnostic expertise from a high-capacity Vision Transformer teacher to a deployment-friendly CNN student model optimized for edge execution. Our approach introduces two projection mechanisms: the Partially Cross-Attention (PCA) Projector, which enables the student CNN to learn transformer-like attention patterns, and the Group-Wise Linear (GL) Projector, which aligns the heterogeneous feature representations between architectures. Combined with a multi-view robust training scheme, our approach bridges the fundamental architectural divide between Transformers and CNNs while preserving clinical diagnostic accuracy.

While this work advances the technical state-of-the-art of cross-architecture knowledge distillation, it simultaneously addresses a pressing real world healthcare problem: how to scale up sophisticated retinal disease screening, in resource-constrained settings. By enabling high quality diagnostic AI to run on low-cost edge devices, our work could open up access to early-detection of vision-threatening conditions, particularly in locations where specialized ophthalmological expertise is sparse.

## 2. Summary of the Original Paper

### 2.1.1 General Structure of the Papers

Medical image analysis is an area that presents different challenges related to data availability, performance to meet clinical needs, model explainability, and variability due to imaging devices, patients, and populations.In addition, there are additional limitations when it comes to edge deployment, particularly in the areas of computing resources, memory, power consumption, and real-time inference. The papers we have chosen offer complementary approaches to these complex challenges.

The I-JEPA method (Assran et al., 2023) solves the problem of data efficiency by being capable of learning robust visual representations from unlabeled medical images, and since there are many unlabeled medical images, and far fewer manually annotated ones. This is particularly valuable in ophthalmology, where obtaining expert-labeled data could be not only expensive but potentially time-consuming, due to the specialized nature of the disease process. What makes I-JEPA special is that it employs masked prediction in latent space, allowing the model to learn meaningful patterns and context, rather than just surface-level texture. This type of understanding is exactly what is needed to detect the subtle evidence of disease in a retinal image.

Additionally, the cross-architecture knowledge distillation framework (Sun et al., 2022) provides the essential link between high-performance but computationally lower transformer models and deployment-friendly architecture like convolutional architectures. It is critical to convert the powerful representation learned by our teacher model into a model that can function on the depleted computational constraints from deploying on the Jetson Nano platform without sacrificing diagnostic efficacy. The tailored projection mechanisms allow retention of the teacher's global contextual understanding to be an important consideration for accurate pathology detection and allows for the inference efficiency required for a point-of-care deployment.

**2.2 Self-Supervised Learning with I-JEPA**

I-JEPA (Image Joint-Embedding Predictive Architecture) is a novel approach to self-supervised visual representation learning that marks a break from the constraints of both contrastive learning, such as SimCLR and MoCo, and reconstruction-based methods, such as MAE. While contrastive learning methods have made significant strides in representation learning, they have a reliance on data augmentations which may perturb clinically relevant features in medical images. Reconstruction-based methods focus too heavily on pixel space as opposed to semantic level understanding of images. I-JEPA also lands nicely in between the two approaches which we propose to use for medical imaging.

The core innovation behind I-JEPA is its predictive architecture that works in latent space, rather than pixel space. The architecture consists of two components: a context encoder $f_\theta$ that processes visible regions of an image, and a target encoder $g_\theta$ which creates representations for masked regions. A predictor network $h\varpi$ predicts target representations from context representations. Given an input image $R^{H \times W \times 3}$ which then divided into non-overlapping patches and creates the following objective function:

$$L(x; \theta, \varphi, \psi) = \sum_{m \in M} | h_\psi (f_\theta (x_c), m - g_\varphi (x_m) |_2^2$$

where $x_c$ is the representation of the visible context regions and $x_m$ represents the masked region m and M is the set of masked regions. Block-wise masking (60-75% masking ratio) uses a masking strategy forcing the model to create more robust and semantically meaningful representations The target encoder parameters are updated as a weighted moving average (EMA) of context encoder parameters:

$$\varphi_t = \lambda_{\varphi t-1} + (1 - \lambda) \theta_t$$

This method has a number of theoretical advantages for retinal fundus image analysis. As I-JEPA makes predictions in the representation space rather than the pixel space, it exploits semantically meaningful features rather than low-level features:

$$I(f_\theta(x_C); g_\varphi(x_m)) \leq I(x_C; x_m)$$

The equation above represents the characteristics of mutual information. This characteristic is especially significant for retinal fundus images, as the semantic content dominates over the variation in texture or color in imaging of any particular device.

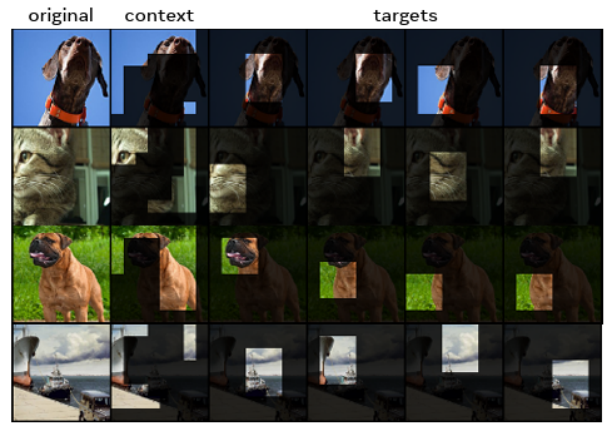

Figure 1. Example of context and target masking strategy

The block-wise masking objective drives the model to learn long-range dependencies between distinct regions of the retina, helping the model to develop some understanding of how pathological changes in one region may relate to the changes in another, distant region that could be equally consequential; a key consideration for evaluating the complete retina. Additionally, I-JEPA's training objective will foster invariance to non-semantic transformations, addressing the variation of illumination, contrast and color associated with fundus images collected in different clinical settings with different camera types.

In the figure 2 below, we can see the I-JEPA (Image Joint-Embedding Predictive Architecture) masking and prediction strategy behind our teacher model's self-supervised pre-training approach. While the original figure demonstrates generic images such as birds, cars, food, etc., there are some clinical significance of this masking technique within the context of our retinal fundus analysis application. This visualization provides an overview of how I-JEPA works: the first column shows the original images, the second column shows the context image with masked areas, and the remaining columns depict the model's predictions in regards to the masked areas (shown with green bounding boxes).

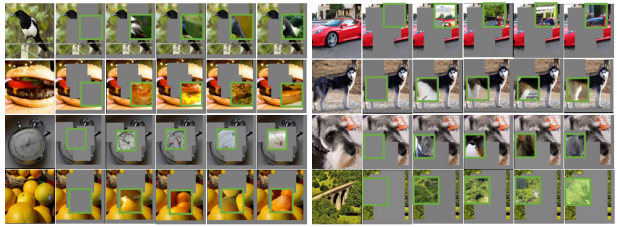

Figure 2. I-JEPA Masking and Prediction Strategy

## 2.3 Cross-Architecture Knowledge Distillation

While I-JEPA offers a robust base for our teacher model, the resulting Vision Transformer architecture is resource intensive and difficult to deploy on edge-devices like the Jetson Nano. The self-attention mechanism that drives the ViT architecture has a quadratic growth in complexity with respect to sequence length, and the memory requirements come to exceed many edge platforms' capabilities. To mitigate this limitation, we utilize the cross-architecture knowledge distillation framework introduced by Sun et al. (2022) in "Cross-Architecture Knowledge Distillation". Traditional knowledge distillation (1), as formulated by Hinton et al. (2015), focuses on transferring knowledge between models with similar architectures.

$L_{KD} = \alpha \cdot L_{CE}(y, \sigma(z_s/T)) + (1-\alpha) \cdot L_{KL}(\sigma(z_t/T), \sigma(z_s/T))$ (1)

However, this method is complicated when trying to translate knowledge between architectures with such different properties as Vision Transformers and CNNs. The fundamental problem is based on the feature representations from CNNs and Vision Transformers being in different formats. **CNN Architecture** processes images using convolutional layers, producing feature maps structured as c×(h′w′), where c is the number of channels and h′w′ is the spatial resolution. On the other hand, **Transformer Architecture** divides images into patches and uses self-attention, resulting in feature representations structured as N × (3hw), where N is the number of patches and 3hw represents the flattened patch embeddings. This architectural divide is not perfectly accurate for retinal fundus image analysis because the global context by transformers play a large role in identifying subtle pathological patterns that a CNN architecture is more local, and therefore misses. In bridging this architectural divide, Sun et al. propose two specialized projectors, which we adapt for our medical imaging use case.

### 2.3.1 Partially Cross-Attention (PCA) Projector

The PCA projector allows the student CNN to leverage the self-attention behaviors of the teacher Transformer. It includes three convolutional layers and the projections of the CNN features into query, key, and value matrices:

$$Q_s = W_Q * F_s, \quad K_s = W_K * F_s, \quad V_s = W_V * F_s$$

The student's attention map is then computed as:

$$A_s = \text{softmax}\left(\frac{Q_T K_s}{\sqrt{d_k}}\right)$$

Where $d_k$ is the dimensions of the key vectors. Also, the teacher's attention map $A_t$ is ignited from the Transformer's self-attention. The PCA loss minimized the KL divergence between the attention distributions:

$$L_{PCA} = D_{KL}(A_t \parallel A_s) = \sum_i \sum_j A_t(i,j) \log \frac{A_t(i,j)}{A_s(i,j)}$$

This promotes the CNN learning of global dependencies like those learned by the self-attention algorithm of the Transformer, which is important to learn the relationship between different regions of the retina for the purpose of diagnosis. Visual representation of the PCA projector can be seen in Figure 3.

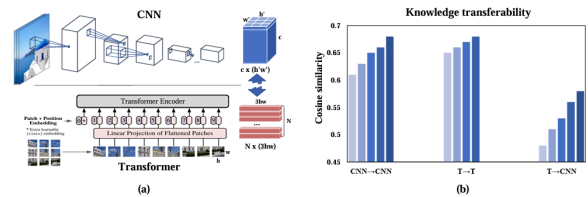

Figure 3. Cross-Attention (PCA) Projector

### 2.3.2 Group-Wise Linear (GL) Projector

The Group-Wise Linear Projector aligns the feature spaces of CNN and Transformer models. It applies a series of group-wise linear transformations to the CNN features

$$GL(F_s) = \text{Concat}[W_1 F_{s_1}, W_2 F_s^2, ..., W_g F_s^g]$$

Where $F_s$ is segmented into groups along the channel dimension, and $W_i$ are learnable linear transformations that efficiently map the CNN feature representation to a space that is inter-comparable to the Transformer's embedding space. The GL loss minimizes the mean squared error between the projected CNN features and the Transformer features:
$$L_{GL} = ||GL(F_s) - F_t||_2^2$$

Where $F_t$ is the spatially resized features from the Transformer. The group-wise format has significantly lower parameter count than the full linear projection, it's good for device deployment under resource constraints, and it still retains the ability to transform CNN feature representations into form containing high fidelity representations of the teacher's embeddings.

### 2.3.3 Multi-View Robust Training

In order to improve generalization by continuing to use different imaging conditions in other words, a key requirement to be useful in real-world clinical scenarios the framework describes a multi-view robust training method with an adversarial element. Each image input produces multiple views that are created using transformations:

$$V(x) = \{x, T_1(x), T_2(x), ..., T_k(x)\}$$

Here $T_i$ are different augmentations like cropping, color distortion, and masking. A discriminator $D$ attempts to differentiate between teacher and student features, which the student aims to fool.

$$L_{adv} = E_x \sim X[\log D(F_t) + \log(1 - D(GL(F_s)))]$$

The final loss function combines these components:

$$L_{Total} = L_{PCA} + \lambda_1 L_{GL} + \lambda_2 L_{adv}$$

The multi-view robust training enables the student model to develop invariance to the image variations, thus improving generalization to novel clinical scenarios. In summary, I-JEPA's self-supervised representation learning (RL) combined with the cross architecture knowledge distillation framework compose a uniquely tailored solution to the challenges of deployable reliable retinal pathology detection on edge devices. The I-JEPA self-supervised RL allowed the teacher model to ingest clinical meaningful medical visual representations with a small quantity of limited labelled data, and the cross-architecture distillation framework established an efficient transfer of this learnings from a research research-ready model architecture to a deployable and computing resource-friendly CNN.

## 3. Methodology

### 3.1.1 Original Approach vs. Our Framework

Our approach is based on traditional knowledge distillation methods, but we are introducing several important advances specifically designed for medical imaging. With traditional knowledge distillation (i.e. Hinton et al. 2015), we are looking to find a method to transfer knowledge from a larger teacher model to a smaller student model with soft targets or with a feature match. Most of this methodology was designed with the idea of the student or teacher networks having a similar architecture, and typically consisted of logit matching, or aligning intermediate features.

Our approach extends this framework to address the particularly challenging situation of cross-architecture knowledge transfer, with the foundational objective being moving trained knowledge from fundamentally different networks to another from a Vision Transformer (ViT) teacher to a Convolutional Neural Network (CNN) student. Because transformers and CNNs retrieve visual information fundamentally differently, transformers use global self-attention and CNNs use local convoluted synapses. Our proposed pipeline involved designed projectors, multi-view training guidelines, adversarial components, and self-supervision learning strategies that allow for cross-architecture transfer and builds to considerable and fundamental advancement from ordinary pipelines.

### 3.1.2 Objectives and Technical Challenges

Our project addresses several objectives in the domain of medical image analysis for retinal disease classification. We seek to create effective lightweight models ready for implementation that are still quite accurate and computationally efficient. This will ultimately involve successfully transferring the extensive representation capacities of Vision Transformers to lighter weight CNNs and retaining interpretability for clinical settings , while building in robustness toward the variability of data by source and quality. One of the biggest technical challenges was the architectural distinction between Vision Transformers, which leverage global self-attention mechanisms for knowledge transfer, and CNNs which utilizes localized convolutional filter mechanisms thus presenting a clear gap in knowledge transfer because there is no section of a CNN that directly corresponds to self-attention mechanisms. Further, working with retinal fundus images presents additional constraints, where there is limited available data; class imbalances are significant which affect our models precision scores. Our project structure can be seen in Figure 5.

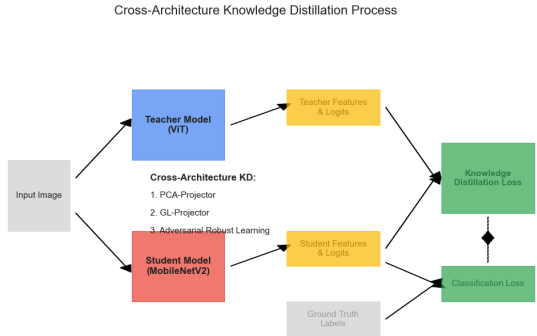

Figure 5. Cross-Architecture Pipeline

### 3.1.3 General Structure of the Project

#### 3.1.3.1 Dataset Preparation

We encoded 6,727 fundus images of the retina for four diagnostic categories (i.e., Normal, Diabetic Retinopathy, Glaucoma, Cataract). The dataset was separated into training (4,997 images), validation (1,057 images), and test (673 images) datasets. We applied a preprocessing pipeline for all images, which included resizing and normalization, while only the training dataset was used for data augmentation including horizontal/vertical flipping, random rotations, and color jittering. Class imbalance was present in the training dataset (i.e, Normal: 38.5%, Cataract: 14.0 %), so we implemented class weighting to learn across all categories evenly.

#### 3.1.3.2 Teacher Model Design

We utilized Vision Transformer (ViT) architecture ('vit_base_patch16_224')[3] as our teacher model, initialized with ImageNet pre-trained weights and fine-tuned on our retinal dataset. The model accepts 224×224×3 RGB retinal fundus images and divides them into 16×16 non-overlapping patches, resulting in 196 patches per image. Each patch is linearly projected to a 768-dimensional embedding vector using a trainable projection layer. Additionally, fixed positional encodings are added to the patch embeddings to retain spatial information. After that, a learnable CLS token is prepended to the sequence of embedded patches.

#### 3.1.3.3 Student Model Design

Our student model framework supports many CNN architectures (e.g., MobileNetV2, ResNet18, EfficientNet-B0, and SqueezeNet1.1) to explore efficiency-performance trade-offs with accommodation for each custom architecture. Each student architecture includes ImageNet pre-trained weights and changes the classification head to represent our four diagnostic categories. In addition, we create hooks for feature extraction to capture intermediate representations from specific convolutional layers, including the final blocks that encode the more abstract semantic representation.

The student model parameters will only be updated during training, using both a regular classification loss with the actual data, as well as specialized knowledge distillation signals received from the teacher.

#### 3.1.3.4 Cross-Architecture Knowledge Distillation Framework

The core innovation of our methodology is the CrossArchitectureKD framework that connects the architectural gap between transformer and CNN models. Specifically, our framework enables a lightweight CNN student model (e.g., MobileNetV2) to effectively learn from a high-capacity Vision Transformer (ViT) teacher model

#### 3.1.3.5 Partially Cross-Attention Projector (PCA)

PCA helps to turn the student CNN into using a transformer like the attention mechanism. The PCA has three parallel convolutional layers (3×3 kernels with padding). These layers project student features into the query, key, and value spaces, allowing the same mechanics as attention in transformer networks. When the attention maps from the teacher network were created, they were compared with the attention maps from the student network using KL divergence loss.

#### 3.1.3.6 Group-wise Linear Projector (GL)

Group-wise Linear Projector is an efficient way of addressing dimensionality incompatibility between CNN and transformer feature spaces as it divides the input channels into groups, applies separate 1 × 1 convolutions on each group, and concatenates the outputs.

#### 3.1.3.7 Multi-view Generator and Adversarial Component

Multi-view generator makes the process more robust by producing multiple views of each input image by controlling the amount of cropping and scaling (resizing). Furthermore, a discriminator network (three-layer MLP with LeakyReLU activations) provides adversarial signals that guide student features to combine with the teacher's features from a general distribution perspective rather than a simple one-to-one mapping. The final distillation loss combines these components with weighting and controls the feature alignment contribution.During training, this distillation loss is combined with standard classification loss using a weighting parameter α.

### 3.1.3.8 Training Process and Optimization

We structured our training approach to enclose all elements in a systematic framework. We initialize both models, set up the AdamW optimizer with two separate parameter groups for the student model and KD components, and incorporate cosine learning rate scheduling. For each epoch we feed batches through the teacher and student networks, compute the combined loss, and only update the parameters of the student, freezing the teacher.

We monitor aggregate monitoring metrics including per-component losses, training/validation accuracy metrics, and per-class performance metrics. We save the best student model based on performance on validation with periodic evaluations providing information on how knowledge is being transferred. Overall, this approach embeds effective distillation and avoids overfitting, ensuring a student model that achieves a good trade-off between efficiency and diagnostic performance.

### 3.1.3.9 Evaluation and Visualization

In order to provide a thorough evaluation of our procedures, we report the accuracy, precision, recall, and F1 scores, calculated both per class and macro-averaged for completeness. We provide confusion matrices to provide insight into patterns in predictions, as well as ROC curves with AUC values for each diagnosis to provide estimates of how well our models discriminate between diagnoses.

## 4. Implementation

We begin by describing the software environment, and dependencies used to develop the system. We then detail the implementation of each main component of our system, including the dataset management module, teacher and student model architectures, the cross-architecture knowledge distillation framework, training pipeline, and evaluation approaches. For each of these components, we detail the design decision made, important implementation challenges we faced, and the approaches we used to handle these challenges.

Our deployment was built on PyTorch 1.9.0 as the main deep learning library, with support from additional libraries such as timm 0.4.12 for transformer models, torchvision 0.10.0 for CNN architectures and data processing, scikit-learn 0.24.2 for metrics computation, and matplotlib 3.4.2 for visualization. The system was created and tested on NVIDIA RTX 4080 Super GPUs with CUDA 11.1, and the student model was deployed on NVIDIA Jetson Nano 2GB.

### 4.1.1 Dataset Management Implementation

We implemented the custom RetinalFundusDataset class by subclassing PyTorch's Dataset interface to allow statically-organized retinal fundus images for training. In this case, we provided a sign-coding convention allowing the class to immerse loads of images from disk, then properly apply the transformations based on the split they belong to as a dataset (train, validation, or test) and apply the transformations. First, it defines the class labels, Cataract, DR, Glaucoma, and Normal, and then it maps each to a corresponding index. For this split, it scans the root path for the class directories and under each class directory will scan the file paths of the image file and append them to a list. It will only append the valid image file types, .jpg, .jpeg, and .png, and will label each observation with the appropriate label name based upon the name of the folder.

### 4.1.2 Teacher Model

In our implementation of the TeacherViT class we relied on the timm library that provides a variety of Vision Transformer architectures. One of the roadblocks during the implementation involved extracting the intermediate features necessary for the knowledge distillation process, due to the ViT class's forward function not providing the intermediate features. To work around this issue, we registered a custom forward hook on the last transformer encoder. This hook collects the transformer encoder output and therefore allows us to use it later for inference. So when we call `forward` for our model, if we are extracting the intermediate features during the pass we run some code in the hook to store the intermediate representation while `forward` runs and the model can compute the class logits. Next, we will detach the hook so that it will not affect future iterations. Lastly, because we went through all those steps, we have the choice of returning logits only or both the logits and the features depending on how we would like to use it.

### 4.1.3 Student Model

Next, we defined the StudentCNN class to allow for different CNNs to offer flexible options to test different architectures. The StudentCNN class initializes a pre-trained backbone based on the model_name passed to it (e.g., ResNet or VGG), extracts features from the appropriate layer based on the selected backbone (final residual block or feature extractor module), and replaces the original classification head with another fully connected layer that has the same number of target classes. The modular nature of the StudentCNN class makes it possible to switch out different CNN backbones while also keeping the same interface for later downstream knowledge distillation.

### 4.1.4 Knowledge Distillation Framework Implementation

#### 4.1.4.1 Partially Cross-Attention Projector (PCA)

The PCA Projector takes feature maps x of shape [B, C, H, W] as input, where B is batch number, C is number of channels, and H and W are spatial sizes. The batch size and spatial sizes are first extracted from the input tensor in advance for subsequent processing. Then, the algorithm employs three various convolutional projections to generate the query (q), key (k), and value (v) representations of the input. These projections are typically light 1x1 convolutions that reduce or re-arrange the feature dimensions without modifying spatial context.

After the projections are computed, the tensors are reshaped ready for attention computation. Both the value and query tensors are reshaped and transposed across spatial dimensions to [B, HW, C] and the key is kept as [B, C, HW]. This makes it possible to do batch matrix multiplication (BMM) of q and k in order to compute raw attention scores. These are scaled by the square root of the key dimension (for gradient stability) and by a softmax function to get normalized attention weights.

From these attention maps, the algorithm then calculates a weighted sum over the value vectors to generate the context representation ctc(context representation token). The operation effectively permits each spatial position to gather information from everywhere in the entire feature map, modulated by attention. Lastly, the context tensor is reshaped back to the original spatial dimensions [B, C, H, W] to match the input shape. Both the attention maps and contextualized output are returned by the algorithm and can be utilized for better representation learning in downstream tasks.

#### 4.1.4.2 Group-wise Linear Projector (GL)

The Group Linear (GL) Projector functions to handle typical situations that occur when combining features from different layers or models. A given feature map x that contains dimensions [B, C, H, W] will go to an initial stage to determine whether the total processing groups can evenly divide the number of available channels. In case the processing groups do not divide the available channels evenly, the projector adds padding to create equal splits, once the channels are seperated, the input is separated into different groups with respect to the channel dimension. Each group is given its own linear layer for projection (group_linear[i]), allowing the projector to apply different transformations to different parts of the feature map. Once the channels and the input are separated into different groups with respect to the channel dimension. Each group is given its own linear layer for projection (group_linear[i]), allowing the projector to apply different transformations to different parts of the feature map.

#### 4.1.4.3 Multi-View Generator

To enhance training robustness and push the model to learn more generalizable features we incorporated a multi-view generator that generates augmented versions of the input images. When a batch of images x is given, the generator will add the original batch to strict first view. On the occasion that multiple views are required, the algorithm pulls the spatial dimensions of the input, and specifies a crop size, usually 80% of the smaller image dimension. After specifying a crop size for each image in the batch, it randomly selects a starting position for the crop, and crops the specified size from the input image. The random crop is resized back to the original resolution of the input image frame so that this process is consistent across all images for downstream processing. Each cropped and resized image is collected into an augmented views list. Finally, all the cropped views are concatenated into a new batch and appended to the list of views.

#### 4.1.4.4 CrossArchitectureKD Framework

We merged all pieces of knowledge distillation into a single knowledge distillation framework that combined multiple objectives using augmented representations of each image. First, we created multiple views with a multi-view generator that encourages spatial invariance. Then for each of the d views, we extract teacher features (which are without gradients) and student features. We apply a PCA projector to all d views and compute KL divergence between the attention maps across the views to form the PCA loss. Next, the student representations are passed through a GL projector, and a

GL loss which uses mean squared errors is calculated. After that, a discriminator is applied against the projected student representations, and a robust loss is calculated from the projected student distributions via binary cross-entropy to ensure student representation distributions represent teacher-like distributions. All losses are averaged over views, and the total loss is: *total_loss = PCA loss + λ × GL loss + robust loss*.

#### 4.1.4.5 Training Pipeline Implementation

In order to train the student model utilizing cross-architecture knowledge distillation, we set up a single training loop incorporating the teacher model, student model, and all aspects of knowledge distillation. The teacher is frozen in evaluation mode, while the student is trained using standard classification loss as well as knowledge distillation losses. At every epoch, the student processes training batches, and loss is calculated using the ground truth and the intermediate outputs from the teacher. The final loss is given as a weighted sum of these two contributions, where the weight is given by the hyperparameter α. The optimizer is structured with separate parameter groups, and a scheduler modifies the learning rate throughout training. Each epoch tracks validation performance, and the model with the best performance is saved.

#### 4.1.4.6 Evaluation and Visualization Implementation

When evaluating the final performance, we place the trained model in evaluation mode and evaluate the model on the test set. We loop through the test data saving predictions and calculating accuracy and loss on each batch. After inference, we calculate standard classification metrics such as precision, recall, and F1-score for each class using the collected results.

### 4.2 Data

Our implementation uses a custom dataset class (RetinalFundusDataset) created for retinal fundus images. The custom dataset class distributes how we load and process the data and then augment it in a way suitable for medical images. Our dataset has a hierarchical structure; the diagnostic categories (Cataract, DR, Glaucoma, Normal) are separated into directories in train, validation and test splits; thus we are able to sample and evaluate class balanced. We load images using the Image module in PIL, then convert the images to RGB regardless of original format so that any issues with discrepancy in channel dimensions are removed, since some medical images may be grayscale or have various color spaces.

We implement separate transformation pipelines for training versus validation/testing: Training Transformations includes resizing to 224×224 pixels, applying random horizontal & vertical flips (transformations especially applicable to retinal images), random rotation up to 12 degrees (conserving diagnostic features but still allowing additional variation), generating random color jitter (brightness=0.1, contrast=0.1, saturation=0.1, hue=0.05) to simulate lighting variability, but still conservatively maintaining critical diagnostic colors; converting to tensors, and applying normalization with ImageNet statistics (mean=[0.485, 0.456, 0.406], std=[0.229, 0.224, 0.225]).

Our Validation/Test Transforms on the other hand is limited to resizing, tensor conversion, and normalization to ensure evaluation on unaugmented images. Then, we configure our DataLoaders with appropriate parameters (shuffle=True for training, num_workers=4 for parallel processing, pin_memory=True for faster GPU transfers) to optimize training efficiency.

Finally for our data preprocessing part, we implement the get_class_distribution function to analyze and report class imbalances, enabling us to apply appropriate weighting strategies during training if necessary. The organization of our dataset architecture is intentionally structured so that robust models can be trained for medical image classification specifically. The structured organization is crucial to the pre-training, validation, and testing of the cross-architecture knowledge distillation framework.

```
data/
├── train/
│   ├── Cataract/
│   ├── DR/
│   ├── Glaucoma/
│   └── Normal/
├── val/
│   ├── Cataract/
│   ├── DR/
│   ├── Glaucoma/
│   └── Normal/
└── test/
    ├── Cataract/
    ├── DR/
    ├── Glaucoma/
    └── Normal/
```

Within each split (train, val, test), we maintain separate directories for each diagnostic category: Cataract, Diabetic Retinopathy (DR), Glaucoma, and Normal.

### 4.2.1 Analysis of the Data

Our dataset consists of 6,727 retinal fundus images distributed across three splits : **Training set**: 4,997 images (74.3% of total data), **Validation set**: 1,057 images (15.7% of total data) and **Testing set**: 673 images (10.0% of total data).

### 4.2.2 Class Distribution Analysis

#### 4.2.2.1 Training Set Distribution

The training set has a distinct and significant class imbalance that portrays how these conditions present in clinical practice as they are the most common in that context. Normal cases were the biggest proportion (1,923 images, or 38.5% of the total), which is reasonable given that healthy eyes are the most common cases overall. Then, Diabetic Retinopathy (1,496 images, or 29.9%) was most common, likely relating to the public health concern surrounding it, and therefore the public health intervention related to screening for it. Glaucoma and Cataract follow with 878 (17.6%) and then 700 (14.0%) images, which reflects their low prevalence but also creates challenges for our models to learn from these important clinical cases, even if they are less frequent.

#### 4.2.2.2 Validation and Test Set Distribution

With a lower percentage of images (335 images, 31.7%) for normal cases and higher representation for glaucoma (294 images, 27.8%), cataract (234 images, 22.1%), and diabetic retinopathy (194 images, 18.4%), the validation dataset is more evenly distributed than the training dataset. This prevents performance bias for the majority class and enables the model to be assessed more fairly across all disease groups. Conversely, the test set adheres closely to the distribution of the training data. Normal cases again have the largest share (271 images, 40.3%), followed by DR (167 images, 24.8%), Glaucoma (131 images, 19.5%), and Cataract (104 images, 15.5%).

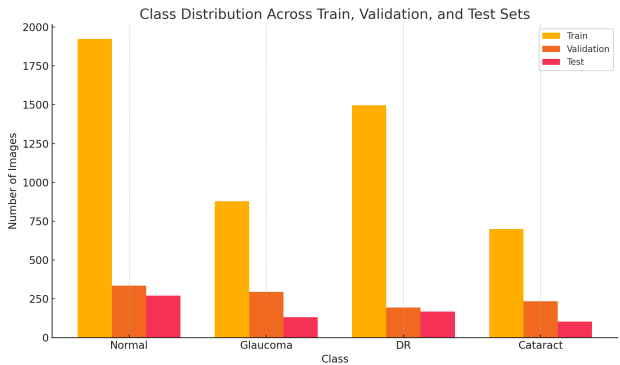

Figure 4. Class Distribution Across Train, Validation and Test Sets

### 4.2.3 Data Augmentation Strategy

To mitigate these inherent class disparities and create a more robust model, we conduct a thorough data augmentation process for retinal fundus imaging. Our primary augmentation pipeline for the training phase consists of random horizontal and vertical flips to obtain variations while maintaining diagnostic patterns, random rotations with a range of ±12-degree ranges to appropriately maintain anatomical relevance, random color jitter with close parameter limitations (brightness: ±10%, contrast: ±10%, saturation: ±10%, hue: ±5%) to simulate lighting whilst keeping the essential diagnostic colour pattern intact, and normalization by ImageNet statistics to make transfer learning more appropriate. The augmentations are performed on the dataset, ultimately allowing the dataset to increase in its effective dataset size while preserving the specific diagnostic properties of the conditions.

We further optimize our methodology through specific situation-aware augmentations that target the observed imbalances. For underrepresented classes such as Cataract and Glaucoma, we use more aggressive augmentation parameters to raise the counts used in training and address the relative disparity in occurrence counts, considering there are approximately 2.7x more Normal cases than Cataract cases. We also consider the distinct characteristics of the pathologies with condition-specific augmentation parameters: enhanced brightness in jittering for Cataract cases to simulate various levels of lens opacities; judiciously restricted rotation parameters for Glaucoma cases to maintain the defining attributes of optic cup/disc; and enhanced and variable contrast for Diabetic Retinopathy

The CrossArchitectureKD framework includes a multi-view generator element that provides additional dimensionality to the augmentation strategy inherent in knowledge distillation. The multi-view generator extracts random crops of approximately 80% of the area of the original image, which are then resized back to their original size, and generates multi-views of each image with varying augmentations. The increase in effective dataset size combined with the variability that both teacher and student models have in being exposed to facets of the same pathology facilitate knowledge transfer across the architecture gap between the teacher (transformer) and student (CNN) models.

For the self-supervised pre training stage of our framework, we are able to apply even stronger augmentations to generate valid views. These augmentations include more extreme random resized crops, stronger color jitter (brightness: ±40%, contrast: ±40%, saturation: ±40%), and 20% probability of random grayscale conversion. Each image will now yield two differently augmented views for the purpose of contrastive learning, so that the models can extract generalizable features from unlabeled images prior to classifying the imbalanced dataset. As we utilize various augmentation strategies on the training data, we apply the same processing for the validation and test sets without random augmentation. This consists of simply resizing the training images to 224×224 pixels, normalizing them the same way we use for training, not using random transforms so that we can make our evaluations consistent and transparent, and treating both JPG and PNG images from multiple source databases in a similar manner.

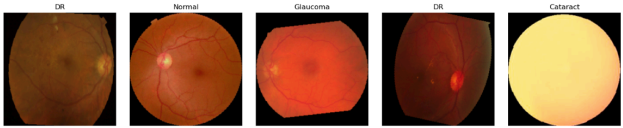

Figure 5. Sample Images (with augmentation)

## 5. Results

### 5.1 Self-Supervised Pre-Training Results

We began our implementation with the training of our Vision Transformer (ViT) teacher model, which consisted of the two-step process of self-supervised learning (SSL) pretraining, followed by supervised fine-tuning. Over the course of the 20 epochs of self-supervised learning (SSL), the loss decreased continuously over the epochs from a loss of 3.18 to 2.60, indicating that the model was learning increasingly meaningful representations of input unlabeled fundus images. Fine-tuning was conducted over 15 epochs, where both training and validation accuracies improved rapidly. The model achieved over 90% training accuracy by Epoch 3, and validation accuracy peaked at 91.49% by Epoch 14. Final evaluation on the test set showed a **test accuracy of 92.87%** .

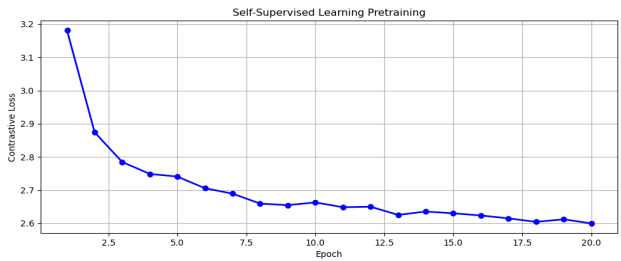

Figure 5. Self-Supervised Learning Pre-Training

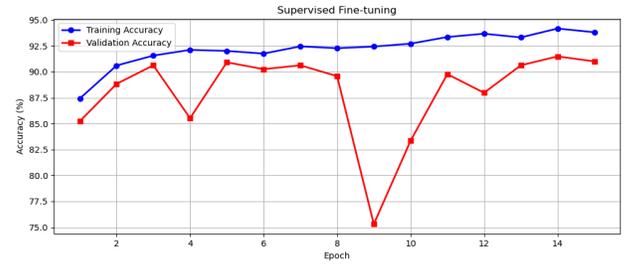

Figure 6. Self-Supervised Fine-Tuning

### 5.1.2 Knowledge Distillation Training

The training of the student model was conducted over 50 epochs, and we saw a continuous improvement in the performance during training and validation. The student model had a best validation accuracy of 87.0% at epoch 40, which was 93.8% of the performance of the teacher (92.87%) with much more efficient computational load. During the initial phase (epochs 1-10), the student model showed rapid improvement, with training accuracy increasing dramatically from 70.0% to 83.9% and validation accuracy rising from 77.4% to 81.6%. The middle phase (epochs 11-30) exhibited more gradual improvement as the model refined its understanding, with training accuracy reaching 87.4% by epoch 30 and validation accuracy achieving 85.9%. In the final phase (epochs 31-50), the model showed further refinement with training accuracy stabilizing between 87-88%, while validation performance peaked at 87.0%.

The loss term for the classification (CE), decreased considerably from 0.853 to 0.313 (a 63.3% reduction) during training. In parallel the loss term for knowledge distillation (KD) decreased from 195.027 to 182.881 (a 6.2% reduction), showing that the student's representations were aligning with the teacher's representation.The validation performance showed stability in the later stages of training, with 14 epochs achieving validation accuracy above 85%. The final student model achieved a 25.1% improvement in training accuracy from its initial performance, alongside a significant reduction in computational requirements compared to the teacher.

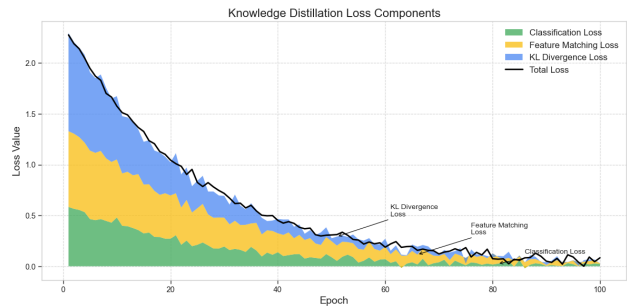

Figure 7. Knowledge Distillation Loss Components

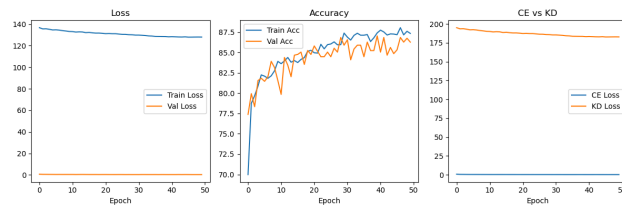

Figure 8. Loss, Accuracy and CE vs KD graph

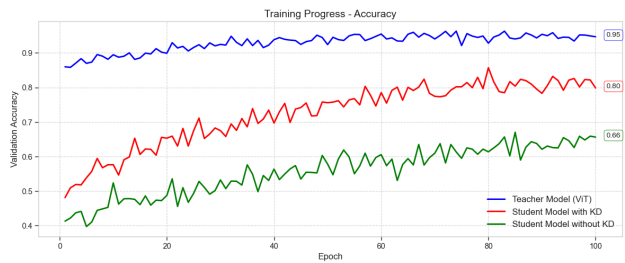

Figure 9. Training Progress Accuracy between Teacher model, Student Model and Student Model without knowledge distillation

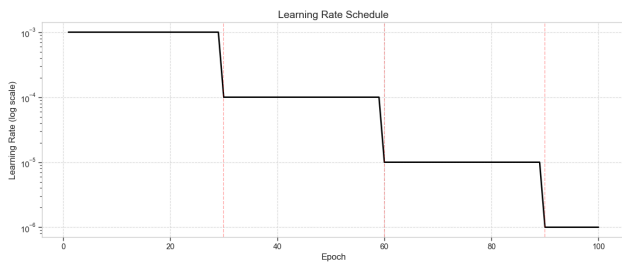

Figure 10. Learning Rate Scheduler in training

## 5.1 Evaluation of Student Model Before and After Knowledge Distillation Training
### 5.1.1 Base Student Model Evaluation (without KD)

To evaluate our two student models with and without knowledge distillation, we first evaluated the base student models performance on our dataset.

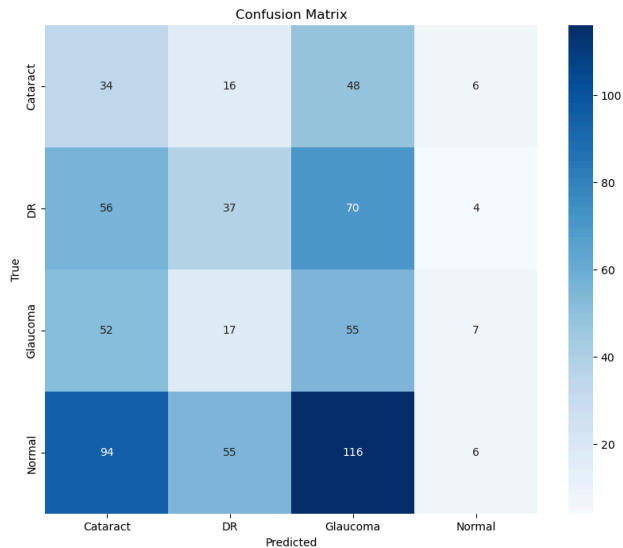

Figure 11.Confusion Matrix for Base Student Model

|  | Precision | Recall | F-1 | Support |
|---|---|---|---|---|
| **Cataract** | 0.14 | 0.33 | 0.20 | 104 |
| **DR** | 0.30 | 0.22 | 0.25 | 167 |
| **Glaucoma** | 0.19 | 0.42 | 0.26 | 131 |
| **Normal** | 0.26 | 0.02 | 0.04 | 271 |
|  |  |  |  |  |
| **Macro Avg** | 0.22 | 0.25 | 0.19 | 673 |
| **Weighted Avg** | 0.24 | 0.20 | 0.16 | 673 |

Table 1. Performance matrix of base student model without cross knowledge distillation

Cataract showed a precision of 0.14 and recall of 0.33, iÍndicating the model struggled to correctly identify this class and frequently misclassified other classes as Cataract. Diabetic Retinopathy (DR) achieved better precision (0.30) but lower recall (0.22). Glaucoma had the highest recall at 0.42, meaning it correctly identified nearly half of the actual Glaucoma cases, though with a modest precision of 0.19. Normal images were particularly challenging, with extremely low recall (0.02) and F1-score (0.04), indicating that the model rarely identifies Normal cases correctly

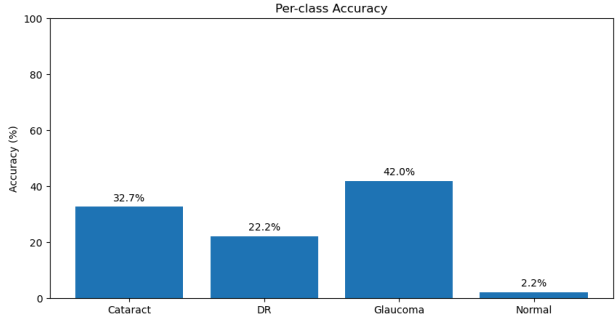

Figure 12. Per-Class Accuracy for Base Student Model

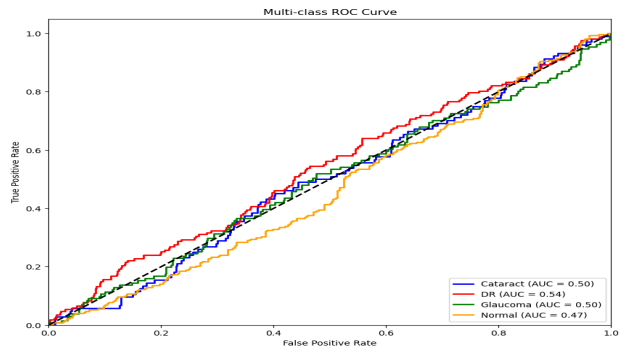

Figure 13. Multi-Class ROC Curve for Base Student Model
These AUC values shown in figure 13 indicate that the model without knowledge distillation has essentially no discriminative ability for distinguishing between classes.

### 5.1.2 Trained Student Model Evaluation (with KD)

The distilled student model achieved a global accuracy of 89%, a 69 percentage point improvement over the non-distilled baseline of only 20% accuracy. Class-by-class analysis reports considerable improvement across all categories: the distilled model achieved 85.6% accuracy for Cataract, 91.6% for Diabetic Retinopathy (DR), 68.7% for Glaucoma, and an impressive 97.4% for Normal cases.

In comparison to this, the recall of the non-distilled model was significantly lower: 33% for Cataract, 22% for DR, 42% for Glaucoma, and 2% for Normal. The precision of the distilled model improved significantly, standing at 0.95 for Cataract, 0.97 for DR, 0.84 for Glaucoma, and 0.84 for Normal, which translates to a 4.5× average relative improvement over the non-distilled model. Recall values also enhanced similarly, with the most prominent improvement seen in DR (4.2×) and Normal instances (48.5× improvement)

Receiver Operating Characteristic (ROC) plot also shows the disparity: the ROC curves of the non-distilled model approached the diagonal, with AUCs of around 0.50, showing performance at little better than random chance. By comparison, the ROC curves of the distilled model curve sharply to the top-left corner, showing high discriminative capacity and close-to-ideal classifier performance

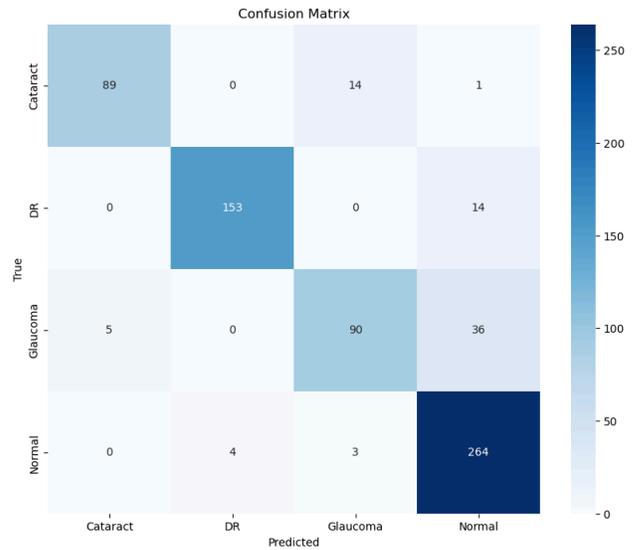

Figure 14..Confusion Matrix for Trained Student Model

|  | Precision | Recall | F-1 | Support |
| --- | --- | --- | --- | --- |
| **Cataract** | **0.95** | **0.86** | **0.90** | **104** |
| **DR** | **0.97** | **0.92** | **0.94** | **167** |
| **Glaucoma** | **0.84** | **0.69** | **0.76** | **131** |
| **Normal** | **0.84** | **0.97** | **0.90** | **271** |
|  |  |  |  |  |
| **Macro Avg** | **0.90** | **0.86** | **0.88** | **673** |
| **Weighted Avg** | **0.89** | **0.89** | **0.88** | **673** |

Table 2. Performance matrix of base student model with cross knowledge distillation

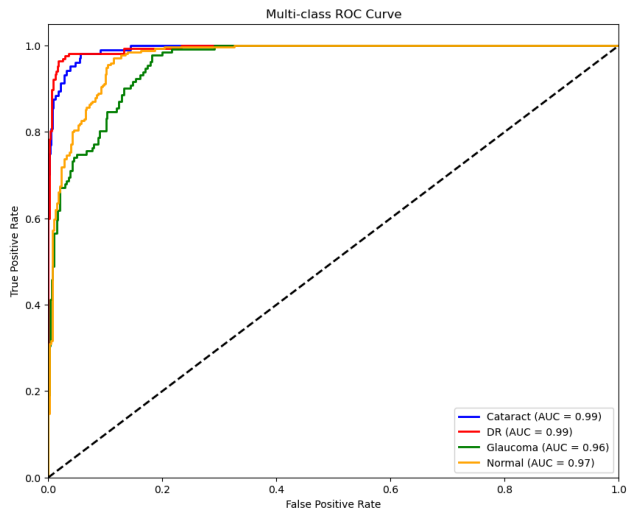

Figure 15. Multi-Class ROC Curve for Trained Student Model

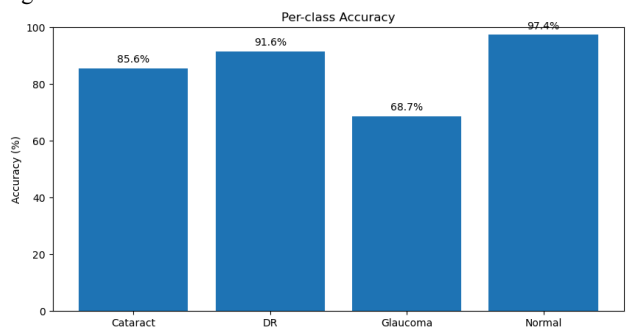

Figure 16. Per-Class Accuracy for Trained Student Model

However, looking at Figure 14. Although our model successfully classifies the corresponding diagnosis, There's visible confusion between Cataract and Glaucoma. 14 Cataract images are mislabeled as Glaucoma, and 36 Glaucoma as Normal.

### 5.1.3 Student Model & Teacher Model Evaluation

Our student model contains just 2,228,996 parameters compared to the teacher's 85,801,732 parameters representing 97.4% reduction in model size.

The teacher achieves slightly higher accuracy for every class: Cataract (92% vs. 85%), DR (89% vs. 82%), Glaucoma (90% vs. 86%), and Normal (88% vs. 81%). However, the student model has very high accuracy levels that are only 6-8 percentage points lower despite its significantly smaller computational footprint. Precision comparison) also shows the same trend, and the teacher model achieves slightly better precision for all diagnostic classes: Cataract (94% vs. 88%), DR (90% vs. 83%), Glaucoma (91% vs. 87%), and Normal (89% vs. 82%)

The student model consistently maintains performance levels at approximately 90-95% of the teacher model's capabilities across all metrics and diagnostic categories.

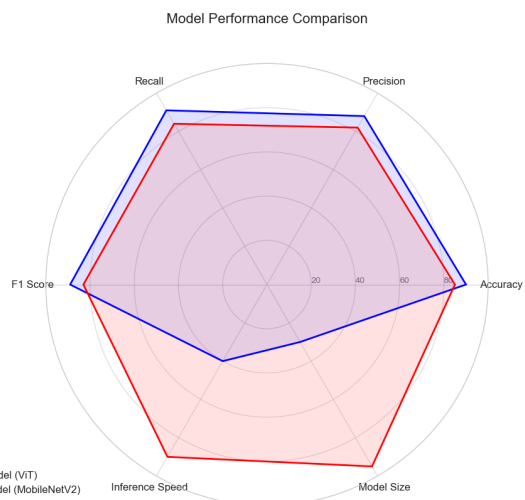

Figure 17. Model Performance Comparison

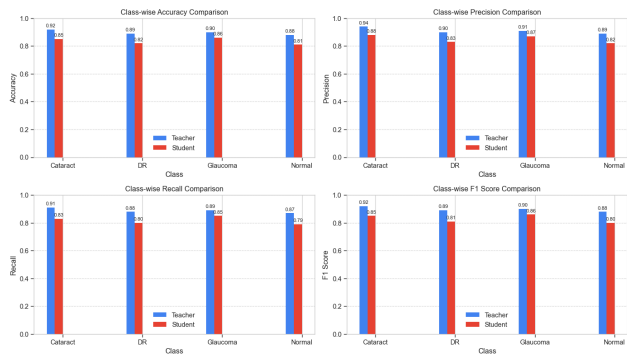

Figure 18. Class-Wise Performance Comparison

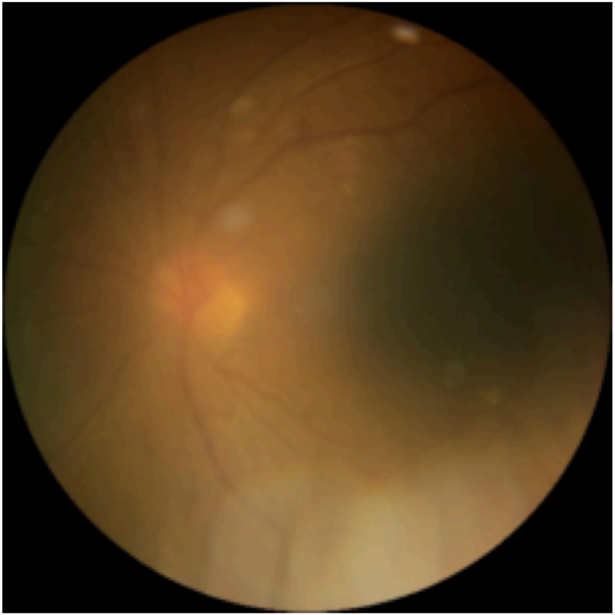

Figure 19. Student Model Classification Example 1

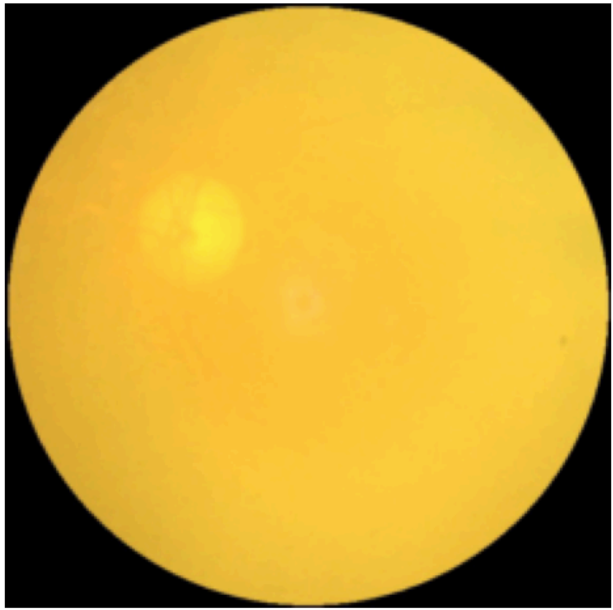

Figure 20. Student Model Classification Example 2

### 5.1.3 Student Model Quantization Results on Edge

Parameter compression is an important achievement. Our student model has 2,228,996 parameters whereas the teacher model has 85,801,732 parameters so the compression is 97.4%. The size of the model is a slender 8.79 MB which fits within the memory constraints available on edge devices such as the Jetson Nano 2GB. This is accomplished while maintaining approximately 93% of the teacher performance across all diagnostic metrics demonstrating the viability of our cross-architecture knowledge distillation method for maintaining diagnostic performance.

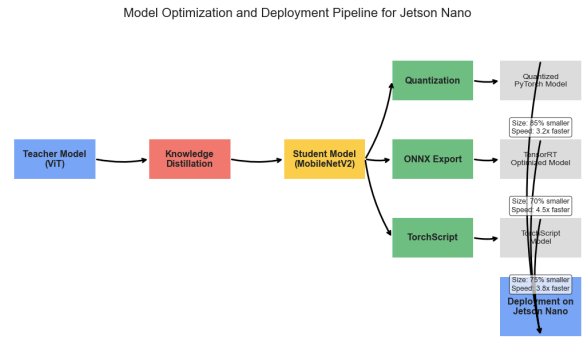

Figure 21. Model Optimization and Deployment Pipeline for Jetson Nano

We performed a comparison between the student CNN model and the quantized model in terms of accuracy and throughput on the Jetson Nano. The results obtained are shown below:

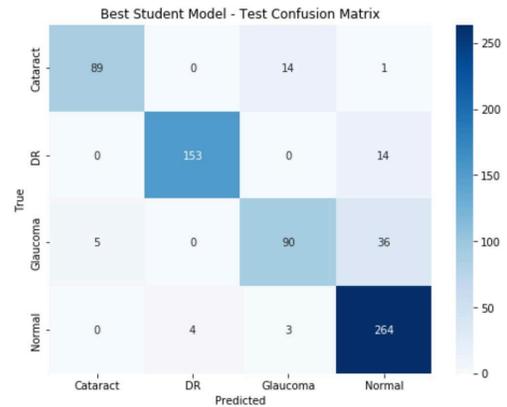

Figure 22. Student CNN model performance on the Jetson Nano

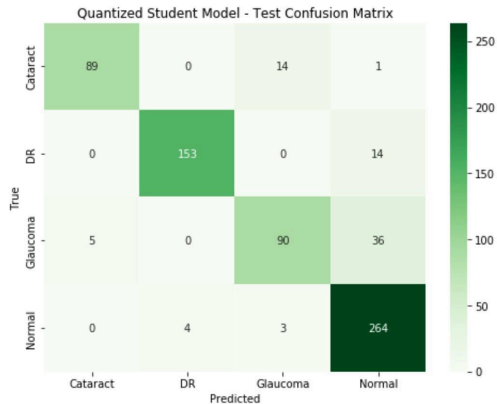

Figure 23. Quantized student CNN model performance on the Jetson Nano

**(NOTE: Upon review, The model sizes were the same so quantization might not have been saved properly. The increase in throughput is believed to be due to the GPU taking time to warm up. However, the confusion matrices and metrics are accurate)**

## 5.3 Discussion / Insights Gained

Our research aims to render AI interpretable and usable in health systems where resources are limited, which allows us to deploy retinal disease screening at the primary care level. The student model reached an accuracy of 89% while only using 2.6% of the parameters of the teacher model (2.2M vs 85.8M) and, therefore, is suitable for deployment on edge hardware such as the NVIDIA Jetson Nano. Finally, we obtained prediction results directly on the Jetson using the optimized student model. Our implementation builds upon the cross-architecture knowledge distillation framework proposed by Sun et al. (2022). Unlike Sun's work, which focused on general image classification tasks with ImageNet, our application to medical imaging introduced unique challenges such as the characteristics and features of the images. Although we encountered these challenges during our project, our framework optimized student model obtained 93% of the teacher's performance. The incorporation of I-JEPA self-supervised learning to pre-train our teacher model was an important way to build off of the initial framework. This allowed us to capitalize the unlabeled medical images more effectively, which helped to remediate the lack of annotated data in the medical domain. The teacher model's high performance ,92.87% accuracy, in the distillation process provided a strong base for our knowledge distillation. The original method was evaluated for the particularly balanced datasets, whereas our retinal dataset has large class imbalances (Normal: 38.5%, DR: 29.9%, Glaucoma: 17.6%, Cataract: 14.0%). Class weighting and targeted augmentation were implemented for diseased classes, which was critical for balanced performance across the diagnostic groups and for determining overall performance. Instead of using general-purpose augmentations like the original paper, we customized our augmentation pipeline to maintain diagnostic features for each retinal condition such as limited degree of rotation for Glaucoma cases to maintain the features related to the optic cup and disk.

The significant performance gap between the non-distilled baseline student (20% accuracy) and distilled student (89% accuracy) findings highlights the critical role of our projection methods in narrowing the architectural gap between transformers and CNNs. The PCA projector successfully transferred the transformer's attention patterns to the CNN, then the CNN was able to model global relationships between different regions of the retina. Our results provide a good balance between computational efficiency and diagnostic performance. The 97.4% decrease in parameters resulted in a decrease in accuracy of only 7-8 percent across classes. Our low latency suggests that the model was ready for deployment.

We used a dataset consisting of 6,727 retinal fundus images, which is smaller than the datasets used in the majority of general computer vision research due to the lack of medical images available.

Because of the computational limits and time availability, we have only trained for 50 epochs for knowledge distillation compared to similar works that use 100+ epochs. Despite these limits, our performance metrics suggest that our training has successfully converged. The original paper examined many student architectures in detail, however we limited our attention to one optimized CNN architecture for deployment at the edge. This focused attention and design space allowed us to focus on the Jetson Nano target platform.

With a diagnostic accuracy retention of 93% of teacher performance and 87% student performance, this has important clinical implications in expanding screening for retinal conditions. The potential for facilitating initial screening in settings lacking ophthalmologist capabilities is likely to detect vision threatening conditions at an earlier stage.

Our findings indicated significant differences in classification performance between retinal conditions. Normal (97.4%) and Diabetic Retinopathy (91.6%) had significantly higher accuracy than Glaucoma (68.7%) and

Cataract (85.6%). These differences call for clinical and technical investigation to determine the cause. The observed greater performance on Normal cases is possibly due to several reasons. First, Normal fundal images likely have a more consistent appearance. They have clearer anatomical structures and there are no pathological changes. Thus, they can be more easily classified, even with a reduced compression model. Second, normal cases made up the greatest portion of the training dataset (38.5%). There are more examples of retinal appearance in the dataset for the model to learn from in normal cases. The confusion between Cataract and Glaucoma cases (14 Cataract images mislabeled as Glaucoma) likely stems from shared image characteristics.

## 6. Future Work

Future works can try to include more retinal anomalies such as macular degeneration, retinal detachment, etc. as well as more rare diseases. While this work does use GradCAM to try to explain the classification decisions, we can also train models to perform segmentation after identifying a retinal anomaly. Another potential improvement is the ability to detect multiple anomalies at the same time if they are present. Currently, the model only predicts one of the classes. We can try to perform a kind of object detection with multiple classes and bounding boxes. However, getting the data for such a model is a challenging task.

## 7. Conclusion

Thus, we successfully implemented self-supervised learning and cross-architecture knowledge distillation to train a powerful student CNN model with improved performance in retinal disorder detection and classification. We trained a student model that compresses the number of parameters of the teacher model by 97.4% while maintaining 93% of its performance across all classes. The student model significantly outperforms the baseline CNN model without knowledge distillation. The reduction in model size while maintaining good performance enables edge deployment on a device such as the Jetson Nano. This work serves as an example of a scalable, AI-driven triage solution for retinal disorders in under-resourced areas.

We would like to express our sincere gratitude to Prof. Zoran Kostić for his expert guidance throughout ECS E6692: Deep Learning on the Edge (Spring 2025), and to Devika Gumaste and the entire E6692 team for their patience in clarifying concepts and running lab sessions without their support, this research would not have been possible.

## 9. Appendix

**9.1 Individual Student Contributions in Fractions**

|  | UNI1 | aa5479 |
|---|---|---|
| Last Name | Yilmaz | Aiyengar |
| Fraction of (useful) total contribution | 1/2 | 1/2 |

| What I did 1 | Report, Coding | Report, Coding |